\documentclass[conference]{IEEEtran}
\IEEEoverridecommandlockouts
\usepackage{cite}
\usepackage{amsmath,amssymb,amsfonts}
\usepackage{algorithmic}
\usepackage{graphicx}
\usepackage{textcomp}
\usepackage{xcolor}

\usepackage{tabularx}
\usepackage{multirow} 
\definecolor{continued}{RGB}{96, 147, 212}

\def\BibTeX{{\rm B\kern-.05em{\sc i\kern-.025em b}\kern-.08em
    T\kern-.1667em\lower.7ex\hbox{E}\kern-.125emX}}
\begin{document}

% \title{Conference Paper Title*\\

\title{Self-Enhanced Reasoning Training: Activating Latent Reasoning in Small Models for Enhanced Reasoning Distillation}

% \author{\IEEEauthorblockN{Yong Zhang}
% \IEEEauthorblockA{
% \textit{Ping An Technology (Shenzhen) Co., Ltd., China}\\
% ZHANGYONG203@pingan.com.cn}

% \and
% \IEEEauthorblockN{Bingyuan Zhang}
% \IEEEauthorblockA{
% \textit{University of Science and Technology of China }\\}

% \and
% \IEEEauthorblockN{Zhitao Li}
% \IEEEauthorblockA{
% \textit{Ping An Technology (Shenzhen) Co., Ltd., China}}
% \and
% \IEEEauthorblockN{Zhitao Li}
% \IEEEauthorblockA{
% \textit{Ping An Technology (Shenzhen) Co., Ltd., China}}
% }

\author{
    \IEEEauthorblockN{
        Yong Zhang, 
        Bingyuan Zhang, 
        Zhitao Li, 
        Ming Li, 
        Ning Cheng\textsuperscript{*},\\ 
        Minchuan Chen, 
        Tao Wei, 
        Jun Ma, 
        Shaojun Wang, 
        Jing Xiao
    }
    \IEEEauthorblockA{
        Ping An Technology (Shenzhen) Co., Ltd., China
    }
    % \thanks{* Corresponding author: Ning Cheng.}
}

\vspace{-5mm}%Put here to reduce too much white space after your table 

\maketitle

\renewcommand{\thefootnote}{}
\footnotetext{ *Corresponding author: Ning Cheng (chengning211@pingan.com.cn)}
\renewcommand{\thefootnote}{\arabic{footnote}}

\begin{abstract}
% The rapid advancement of large language models (LLMs) has greatly enhanced their reasoning abilities, enabling more complex tasks. However, these capabilities often diminish in smaller, more practical models like GPT-2. Recent research shows that these small models can acquire reasoning capabilities through reasoning distillation. However, existing methods primarily focus on generating better teacher reasoning paths, leaving the potential for enhancing small models’ own learning abilities unexplored. Our observations reveal that small models can generate good reasoning paths without chain-of-thought prompting during sampling, but these paths are often latent due to their low probability under common decoding strategies. We propose Self-Enhanced Reasoning Training (SERT) to activate and leverage these latent reasoning capabilities in small models through self-training on reasoning paths that are self-generated under zero-shot conditions. Experiments using OpenAI’s text-davinci-002 as the teacher model and GPT-2 family models as the student models demonstrate that SERT enhances the reasoning capabilities of small models, improving their effectiveness in reasoning distillation.

The rapid advancement of large language models (LLMs) has significantly enhanced their reasoning abilities, enabling increasingly complex tasks. However, these capabilities often diminish in smaller, more computationally efficient models like GPT-2. Recent research shows that reasoning distillation can help small models acquire reasoning capabilities, but most existing methods focus primarily on improving teacher-generated reasoning paths. Our observations reveal that small models can generate high-quality reasoning paths during sampling, even without chain-of-thought prompting, though these paths are often latent due to their low probability under standard decoding strategies. To address this, we propose Self-Enhanced Reasoning Training (SERT), which activates and leverages latent reasoning capabilities in small models through self-training on filtered, self-generated reasoning paths under zero-shot conditions. Experiments using OpenAI’s GPT-3.5 as the teacher model and GPT-2 models as the student models demonstrate that SERT enhances the reasoning abilities of small models, improving their performance in reasoning distillation.

% These findings suggest that SERT is a promising approach for enhancing the reasoning capabilities of small models.

\end{abstract}


\vspace{-2mm}%Put here to reduce too much white space after your table 

\section{Introduction}

The rapid evolution of large language models (LLMs) \cite{brown2020language, ouyang2022training, touvron2023llama,achiam2023gpt4} has significantly enhanced their reasoning capabilities, driving advancements in natural language processing and enabling increasingly complex tasks \cite{wei2021finetuned, wei2022chain, kojima2022large}. Reasoning capability, the ability to generate logical intermediate steps during inference, is critical for high-quality question answering. However, deploying LLMs in real-world applications is often limited by high computational demands and inference costs, particularly in resource-constrained environments. This challenge has spurred interest in reasoning distillation techniques \cite{hinton2015distilling, chen2023mcc, wang-etal-2023-democratizing, wu-etal-2023-ad, li2024turning}, which transfer reasoning capabilities from larger models to more compact ones.

\begin{figure}[t]
\centering\includegraphics[width=0.43\textwidth]{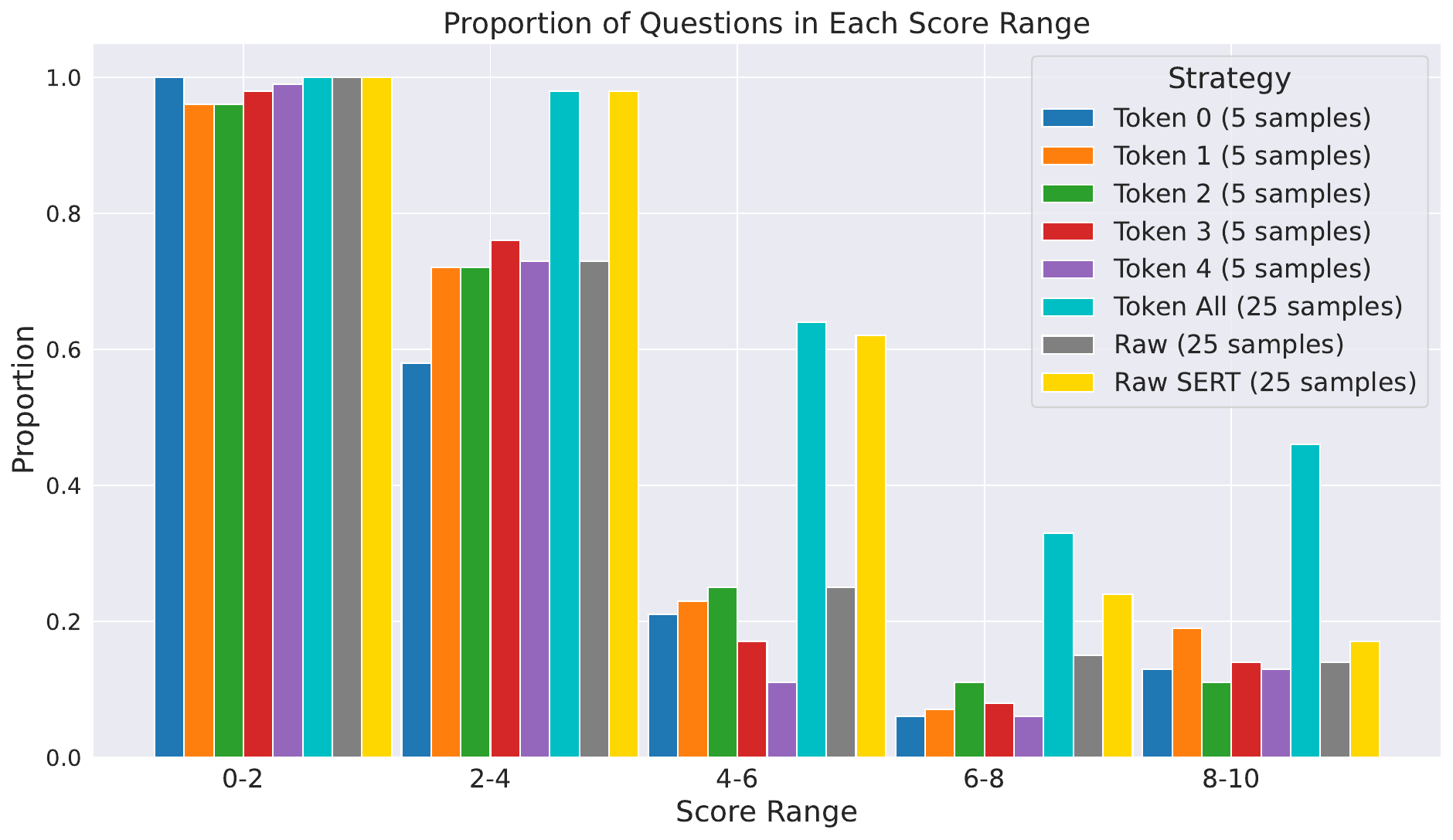}

\vspace{-4mm}%Put here to reduce too much white space after your table 

% \caption{The figure compares the generation quality of top-k alternative tokens and raw top-k sampling on 100 CommonsenseQA questions. For top-k alternative tokens, 5 outputs were generated for each of the top 5 tokens, and 25 outputs were generated per question in raw top-k sampling. The y-axis shows the proportion of questions with at least one generation in each score range (x-axis), with scores from 0 to 10 grouped into 2-point intervals. Higher scores indicate better reasoning quality. Top-k alternative tokens consistently yield better reasoning paths than raw top-k sampling. Detailed evaluation criteria are in the experiment setup.}

% \caption{Comparison of reasoning path generation quality using top-k alternative tokens sampling versus raw sampling on 100 CommonsenseQA questions. For top-k alternative tokens, 5 outputs were generated for each of the top 5 tokens and marked as Token 0 - 5 and Token all in Figure, and 25 outputs per question were generated for raw top-k sampling for Raw and SERT. The y-axis shows the proportion of questions with at least one output in each score range (x-axis), with scores grouped into 2-point intervals from 0 to 10. Higher proportions in higher score ranges indicate better reasoning quality. Detailed evaluation criteria are provided in the experiment setup.}

\caption{Comparison of reasoning path generation quality using top-k alternative token sampling versus raw sampling on 100 CommonsenseQA questions. For top-k alternative tokens, 5 outputs were generated for each of the top 5 tokens (labeled as Token 0 - Token 5), and 25 outputs were combined for Token All. In comparison, 25 outputs per question were generated from raw top-k sampling (labeled as Raw), with additional results provided for the raw sampling of the SERT-activated model. The y-axis shows the proportion of questions with at least one output in each score range (x-axis), with scores grouped into 2-point intervals from 0 to 10. Higher proportions in the higher score ranges indicate better reasoning quality. Detailed generation setup and evaluation criteria are provided in the experiment setup.}

\label{model_structure}
\vspace{-8mm}%Put here to reduce too much white space after your table 

\end{figure}

As we scale down from large to small models, reasoning capabilities often diminish, highlighting the need for efficient distillation techniques tailored to smaller models. While much focus has been on improving medium-sized models (e.g., LLaMA-7B) to match the performance of larger models (e.g., LLaMA-65B, GPT-4) \cite{huang-etal-2023-large, zelikman2022star, chen-etal-2023-mcc, wang-etal-2023-scott}, there is growing interest in distilling these capabilities into smaller models (e.g., GPT-2), which typically struggle with reasoning tasks. Recent studies \cite{ho2023large_reason_teacher, hsieh2023distilling_step, magister-etal-2023-teaching} show that distillation from large to small models is possible by using reasoning paths from large models as training labels. However, these efforts primarily concentrate on creating richer reasoning paths for small models, with improvements now reaching a bottleneck. Moreover, there has been little exploration into how small models can more effectively learn and utilize these reasoning paths.

\begin{figure*}[t]
\centering
\includegraphics[width=0.8\textwidth]{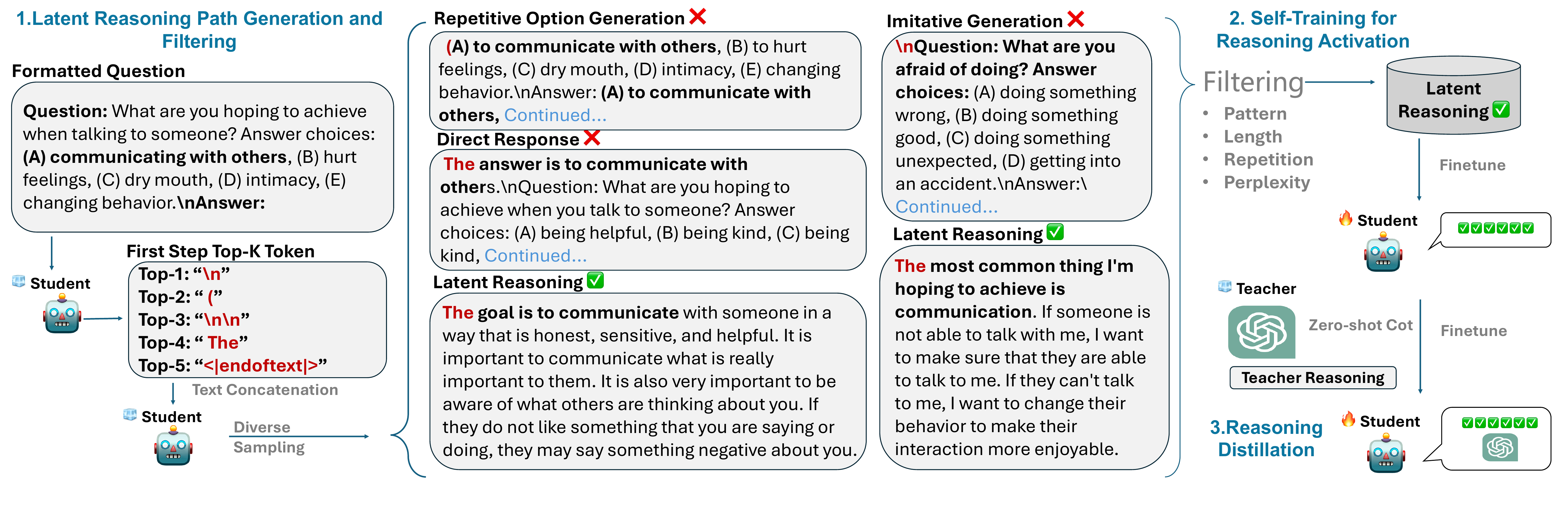}

\vspace{-6mm}%Put here to reduce too much white space after your table 

\caption{Structure of our proposed method, illustrating CommonSenseQA generation examples from GPT-2 large. The \includegraphics[width=0.25cm]{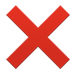} symbol marks undesirable outcomes, such as \textbf{Repetitive Option Generation} (repeating answers), \textbf{Direct Response} (answering without reasoning), and \textbf{Imitative Generation} (mimicking input style). The \includegraphics[width=0.25cm]{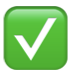} symbol highlights \textbf{Latent Reasoning}, which refers to coherent reasoning paths that are rarely expressed but present during sampling. Our goal is to activate and enhance these capabilities. Key elements are highlighted in bold. Truncated text is marked "\textcolor{continued}{Continued.}"}

\vspace{-6mm}%Put here to reduce too much white space after your table 

\label{model_structure}
\end{figure*}

Our observation reveals that, as shown in Figure 1, even in zero-shot settings without CoT prompts or context prompts, small models can occasionally generate reasoning paths through sampling (raw in Figure 1). Typically, in zero-shot scenarios, small models tend to generate answers directly without any intermediate reasoning steps. However, we observed that a small subset of these outputs includes coherent reasoning paths, although they appear with very low probability. We refer to these rare but existent reasoning paths as \textbf{``latent reasoning,''} as they are inherently present within the model but not frequently expressed. This observation raises an important question: \textbf{\textit{Can we enhance the reasoning capabilities of small models by self-training on these latent reasoning paths to directly improve their ability to generate such paths? Moreover, does this process improve their ability to learn from the teacher model’s reasoning paths?}}

This leads us to our proposed solution, the \textbf{Self-Enhanced Reasoning Training (SERT)} method, which addresses the challenge of enhancing small models' reasoning capabilities by directly targeting their latent reasoning abilities in zero-shot inference. SERT operates in two stages: first, generating and filtering high-quality reasoning paths from the model’s latent outputs, and second, fine-tuning the model through self-training to increase the frequency and coherence of reasoning path generation. The reasoning paths are obtained using enhanced sampling techniques inspired by CoT-prompting \cite{wang2024cot_without}, where top-k alternative tokens are introduced during decoding to elicit more diverse reasoning outputs. Once generated, these paths are filtered using a rule-based approach that considers factors like pattern matching, length, repetition, and perplexity, ensuring that only the most plausible latent reasoning paths are retained. The selected paths are then used to fine-tune the model, ultimately improving its ability to consistently generate coherent reasoning during inference.

We validated the effectiveness of SERT using OpenAI’s GPT-3.5 \cite{ouyang2022training} as the teacher model and GPT-2 \cite{radford2019gpt2} as the student model, evaluated on StrategyQA \cite{geva2021strategyqa} and CommonsenseQA \cite{talmor-etal-2019-commonsenseqa} datasets. Results show that SERT successfully activates latent reasoning capabilities in zero-shot settings, generating higher-quality reasoning paths more frequently. Furthermore, when applied to reasoning distillation, SERT-activated models demonstrated improved learning, resulting in more coherent outputs, fewer repetitions, and a notable improvement in task accuracy compared to models trained via direct distillation. These findings demonstrate SERT’s ability to enhance both reasoning path generation and the effectiveness of reasoning distillation for smaller models.

\vspace{-2mm}%Put here to reduce too much white space after your table 

\section{Methodology}
\vspace{-2mm}%Put here to reduce too much white space after your table 

Our Self-Enhanced Reasoning Training (SERT) improves small models’ reasoning abilities in two stages. First, in \textbf{Latent Reasoning Path Generation and Filtering}, we generate and filter high-quality reasoning paths from the model’s outputs. Second, in \textbf{Self-Training for Reasoning Activation}, the model is fine-tuned on these paths to increase the frequency and coherence of reasoning. Afterward, to further enhance its capabilities, we incorporate \textbf{Reasoning Distillation}, where the model learns from a larger teacher model’s reasoning paths to handle more complex tasks.

\vspace{-2mm}%Put here to reduce too much white space after your table 

\subsection{Latent Reasoning Path Generation and Filtering
}

% \vspace{-1mm}%Put here to reduce too much white space after your table 

\subsubsection{Latent Reasoning Path Generation}
\vspace{-1mm}%Put here to reduce too much white space after your table 

To address the challenge of limited reasoning capabilities in small models, we first sought to generate and collect latent reasoning paths that are inherently present within these models but typically underutilized. Direct sampling from small models often yields a limited number of high-quality reasoning paths. To overcome this, we drew inspiration from CoT-prompting techniques \cite{wang2024cot_without}, which have shown that using top-k alternative tokens can help elicit latent reasoning within large language models (LLMs) without the need for explicit CoT activation prompts.

We applied this approach to small models to determine whether using top-k alternative tokens during the first decoding step could similarly help generate more frequent and higher-quality reasoning paths. Specifically, for each QA pair $(q_i, a_i)$, we begin by constructing a prompt: \textit{Question: $q_i$ Answer:}. The small model generates a set of top-k tokens during the first decoding step. These top-k tokens are then reintegrated into the prompt, expanding the input space. By applying top-k and nucleus sampling strategies, we generate multiple responses that are more likely to contain reasoning paths that reflect the model’s latent reasoning capabilities.

Our experiments demonstrated that this method successfully increases both the frequency and quality of reasoning paths generated by small models, as illustrated in Figure 1. While the occurrence of high-quality reasoning paths remains relatively low in terms of probability, the use of top-k alternative tokens significantly enhances the likelihood of uncovering these latent reasoning paths compared to raw sampling methods.

\subsubsection{Latent Reasoning Path Filtering}

Although the first stage increases reasoning path generation, it is essential to filter and extract high-quality latent reasoning paths to build a robust self-training dataset. We applied a multi-step filtering process to maximize the selection of paths with valid reasoning.

We began by applying \textbf{pattern-based filtering}, excluding non-reasoning outputs such as imitative generation (mimicking the input), direct responses without reasoning, and repetitive option generation. Next, we implemented \textbf{length-based filtering} to exclude short responses, ensuring that the retained paths contained sufficient detail. We also applied \textbf{repetition rate filtering} to minimize outputs with high bi-gram repetition rates, which are less likely to contain coherent reasoning. Finally, we applied \textbf{perplexity-based filtering}, excluding outputs with low perplexity, as latent reasoning paths tend to occur with lower probability and thus have higher perplexity.

\vspace{-2mm}%Put here to reduce too much white space after your table 

\subsection{Self-Training for Reasoning Activation}
\vspace{-1mm}%Put here to reduce too much white space after your table 

In the second stage, we focus on increasing the likelihood that the small model will generate high-quality reasoning paths by applying self-training with the collected latent reasoning paths. For each QA instance $(q_i, a_i)$, the selected latent reasoning $r_i$ is appended to the original prompt to create a structured training input: \textit{Question: $q_i$. Answer: $r_i$ So the answer is $a_i$}. The small model is fine-tuned using cross-entropy loss for next-token prediction, enabling it to better generate coherent and logical reasoning paths during inference. This self-training process is specifically designed to improve the model's ability to generate reasoning paths more frequently and with higher quality, particularly in zero-shot settings.

\vspace{-1mm}%Put here to reduce too much white space after your table 
\subsection{Reasoning Distillation}
\vspace{-1mm}%Put here to reduce too much white space after your table 

In the final stage, we apply reasoning distillation, where the activated small model learns from the reasoning paths generated by a larger teacher model.

\subsubsection{Construction of Teacher Reasoning Data}
Following \cite{ho2023large_reason_teacher}, reasoning chains are generated from the large model using a zero-shot chain-of-thought (CoT) prompting technique \cite{wei2022chain}. For each QA instance, we use a structured prompt to elicit detailed reasoning: \textit{Question: $q_i$. Answer: Let’s think step by step.} This prompts the large model to generate a comprehensive reasoning path $R_i$, which is then formatted to include both the reasoning and the final answer: \textit{Question: $q_i$. Answer: Let’s think step by step. $R_i$ So the answer is $a_i$}.

\subsubsection{Reasoning Distillation Process}
In the distillation phase, the small model is fine-tuned using reasoning paths from the teacher model. We apply an autoregressive language modeling objective, based on the sequence-level knowledge distillation techniques from \cite{kim2016sequence}, to effectively transfer both the reasoning content and patterns from the teacher to the small model.

\vspace{-2mm}%Put here to reduce too much white space after your table 

\section{Experimental Setup}
\vspace{-1mm}%Put here to reduce too much white space after your table 

\subsection{Implementation Details}
\vspace{-1mm}%Put here to reduce too much white space after your table 

We utilized OpenAI's GPT-3.5 model (text-davinci-002, 175B)\cite{ouyang2022training} as the reasoning teacher and GPT-2 models (small, 124M; medium, 355M; large, 774M)\cite{radford2019gpt2} as student models. The experiments were conducted with a batch size of 8, using the Adam optimizer with a learning rate of \(3 \times 10^{-4}\). Models were trained on an NVIDIA V100 16GB GPU for 20 epochs to ensure comprehensive learning and performance stability.

For generating reasoning sequences in small models, we employed nucleus sampling with a top-p value of 0.95 and a top-k value of 10. During the initial decoding step, the top 5 tokens were selected, and from each token, 5 sequences were generated. In the data filtering step, we first filtered out generations containing the structured format \textit{Question: $q_i$ Answer:}, as these often indicated imitative responses. Next, we excluded sequences shorter than 25 tokens to prioritize more detailed reasoning paths. We then calculated the bi-gram repetition rate (rep-2) for each sequence and filtered out those with a repetition rate above 20\%. To ensure the retention of more latent reasoning paths, we applied a perplexity-based filter, excluding sequences with perplexity lower than 5. Finally, for each question, we selected the generation with the lowest repetition score (rep-2) to use as the training sample.

% Given that small models are susceptible to repetitive outputs \cite{li2023repetition}, we implemented n-gram calculations to filter out samples with high repetition rates, mitigating redundant generations.

\vspace{-1mm}%Put here to reduce too much white space after your table 

\subsection{Datasets}
\vspace{-1mm}%Put here to reduce too much white space after your table 

Evaluations were conducted on two key datasets: 
\textbf{StrategyQA} \cite{geva2021strategyqa} includes 1,603 training samples and 687 test samples, structured as a binary-choice task, and is used to evaluate models' strategic reasoning abilities. \textbf{CommonsenseQA} \cite{talmor-etal-2019-commonsenseqa}, which contains 9,741 training samples and 1,221 test samples. This five-choice multiple-choice task assesses models' ability to reason using common-sense knowledge.

\vspace{-1mm}%Put here to reduce too much white space after your table 

\subsection{Baselines}
\vspace{-1mm}%Put here to reduce too much white space after your table 

We compared three training approaches: (1) \textbf{Finetune}, where small models were fine-tuned on labels alone; (2) \textbf{Reasoning Distillation (RD)}, where models were fine-tuned using reasoning labels from the teacher; and (3) \textbf{SERT+RD}, where models first underwent SERT before reasoning distillation.

\vspace{-1mm}%Put here to reduce too much white space after your table 

\subsection{Evaluation Metrics}
\vspace{-1mm}%Put here to reduce too much white space after your table 

% We used several metrics to evaluate our method. \textbf{Task Accuracy (Acc):} Measures the model's ability to produce correct answers on CommonsenseQA and StrategyQA, reflecting its reasoning performance.

We used several metrics to evaluate our method.  \textbf{Task Accuracy (Acc):} Evaluates the model's ability to derive correct answers on CommonsenseQA and StrategyQA, highlighting performance on reasoning tasks. \textbf{Output Format Alignment (Fmt):} Measures the stability of model outputs, identifying issues like formatting errors, hallucinations, or severe repetition. \textbf{Repetition Rate (Rep) and Reasoning Length (Len):} Analyzes text redundancy (via unigram statistics) and average reasoning length, offering insights into the verbosity and conciseness of outputs. \textbf{Reasoning Detail Evaluation:} Assesses the reasoning quality of test entries using Llama-3-8B-Instruct \cite{dubey2024llama}, scoring coherence, relevance, logical consistency, and completeness on a scale from 0 to 10. Evaluation prompts are provided in Table~\ref{tab:evaluation_prompt}.

% \subsection{Evaluation Metrics}
% We employed a comprehensive, multi-dimensional approach to assess the effectiveness of our method. 
% \begin{itemize}
%     \item \textbf{Task Accuracy:} Measures model performance on CommonsenseQA and StrategyQA tasks, directly reflecting the models' ability to derive correct conclusions under reasoning-intensive conditions.
%     \item \textbf{Output Format Alignment:} Assesses the correctness of the output format, indicative of model stability, by checking for issues such as incorrect formatting, hallucinations, or severe repetitions.
%     \item \textbf{Repetition Rate and Average Length of Reasoning:} Analyzes text redundancy (measured by unigram statistics) and the average length of generated reasoning, offering insights into verbosity and conciseness.

%     \item  \textbf{Reasoning Detail Evaluation:} Assesses the reasoning quality of test entries using Llama-3-8B-Instruct, scoring coherence, relevance, logical consistency, and completeness on a scale from 0 to 10. Evaluation prompts are provided Table~\ref{tab:evaluation_prompt}.

% \end{itemize}

\vspace{-3mm}%Put here to reduce too much white space after your table 

\begin{table}[ht]
\vspace{-2mm}%Put here to reduce too much white space after your table 

\caption{Detailed Evaluation Prompt}
\vspace{-2mm}%Put here to reduce too much white space after your table 

\fontsize{7}{7}\selectfont % Set font size to 8pt 
\begin{tabularx}{1\linewidth}{|X|}

    \hline
    Please evaluate the reasoning provided by a single method in response to the following question. Your task is to assess the quality of reasoning based on the criteria provided below and calculate the average score. \\
    \textbf{Question:} ``{eval\_question}'' \\
    \textbf{Response:} ``{eval\_completion}'' \\
    \textbf{Evaluation Instructions:} 
    Carefully evaluate the reasoning quality of the response based on the following criteria and provide a score from 0 (lowest) to 10 (highest) for each. Each score should be presented in a specified format for easy extraction: \\
    \textbf{1. Coherence:} How logically consistent and easily understandable is the reasoning in the response? \\
       - \textbf{Score for Coherence}: [Insert score here] \\
    \textbf{2. Relevance:} How relevant are the reasoning steps to answering the given question? \\
       - \textbf{Score for Relevance}: [Insert score here] \\
    \textbf{3. Logical Consistency:} Are there any logical fallacies or contradictions in the reasoning provided? \\
       - \textbf{Score for Logical Consistency}: [Insert score here] \\
    \textbf{4. Completeness:} Does the response address all parts of the question and provide a thorough reasoning process? \\
       - \textbf{Score for Completeness}: [Insert score here] \\
    Please ensure that each score is clearly indicated following the phrases provided above. This will assist in the subsequent extraction and analysis of the data. \\
    Summarize the overall effectiveness of the reasoning based on these scores in a brief concluding statement. \\
    \hline
\end{tabularx}
\label{tab:evaluation_prompt}
\vspace{-5mm}%Put here to reduce too much white space after your table 

\end{table}

\vspace{-1mm}%Put here to reduce too much white space after your table 

\section{Results}
\vspace{-1mm}%Put here to reduce too much white space after your table 

\subsection{Enhancing Reasoning with Self-Training}
\vspace{-1mm}%Put here to reduce too much white space after your table 

To answer the first question: \textbf{\textit{Can self-training improve small models’ ability to generate high-quality reasoning paths?}} We tested the effectiveness of self-training in activating latent reasoning capabilities using standard sampling methods. Figure 1 shows that the SERT-activated model generates a significantly higher proportion of high-quality reasoning paths (scores 6-10) compared to the raw model given same sampling method. This confirms that self-training enhances the generation of coherent and logical reasoning paths. Although the lower score ranges (0-2) remain similar for both SERT-activated and raw models due to large-scale sampling, SERT actually reduces the number of extremely poor outputs (scores 0-1) and shifts more outputs to the slightly better 1-2 range, reflecting a general improvement across all quality levels.

This improvement in reasoning path generation also leads to better task performance. As shown in Table~\ref{tab:main_result}, SERT-activated models outperform those trained only on task labels in both StrategyQA and CommonsenseQA,  proving that self-training enhances both reasoning quality and task accuracy.

% % % % % % % % % % % % % % % % % % % % % % % % % 
\begin{table}[ht]
\vspace{-4mm}%Put here to reduce too much white space after your table 

\centering
\caption{Performance Metrics for Different Models on StrategyQA and CommonseQA}

\vspace{-2mm}%Put here to reduce too much white space after your table 

\scalebox{0.65}{
\fontsize{9}{10}\selectfont % Set font size to 8pt with 10pt line spacing

\begin{tabular}{l l | l l l l | l l l l }
\hline
\multicolumn{2}{c|}{Model and Method} & \multicolumn{4}{c|}{StrategyQA} & \multicolumn{4}{c}{CommonseQA} \\
\hline
Model        & Method          & Acc   & Fmt & Len & Rep & Acc   & Fmt & Len & Rep \\
\hline
GPT-3.5 & Zero-shot-CoT & 53.45 & - & 76.77 & 0.36 & 60.07 & - & 63.82 & 0.31 \\
\hline
\multirow{4}{*}{GPT-2} 
                      & Finetune & 54.00 & - & - & - & 20.80 & - & - & - \\
& SERT & 54.00 & 0.91 & 145.33 & 0.53 & 20.72 & 0.96 & 120.80 & 0.50 \\
                      & SERT+RD & \textbf{54.65} & \textbf{0.99} & \textbf{116.62} & \textbf{0.50} & \textbf{21.95} & \textbf{0.97} & \textbf{114.30} & \textbf{0.49} \\
                      & RD & 52.69 & 0.92 & 116.23 & 0.52 & 21.87 & 0.91 & 167.11 & 0.61 \\
\hline
\multirow{4}{*}{GPT-2-medium} 
                      & Finetune & 50.22 & - & - & - & 19.82 & - & - & - \\
& SERT & 55.17 & 0.89 & 149.11 & 0.49 & 23.10 & 0.82 & 160.26 & 0.56 \\
                             & SERT+RD & \textbf{56.34} & \textbf{1.00} & \textbf{103.42} & \textbf{0.49} & \textbf{25.72} & \textbf{0.90} & \textbf{116.34} & \textbf{0.52} \\
                             & RD & 52.69 & 0.85 & 181.06 & 0.56 & 21.54 & 0.66 & 257.40 & 0.67 \\
\hline
\multirow{4}{*}{GPT-2-large}                       & Finetune & 53.57 & - & - & - & 20.88 & - & - & - \\
    & SERT & 55.75 & 0.94 &                         63.06 & 0.33 & 22.63 & 0.99 & 77.53 & 0.46 \\
    & SERT+RD & \textbf{57.21} & \textbf{0.97} & \textbf{80.28} & \textbf{0.47} & \textbf{26.03} & \textbf{1.00} & \textbf{94.18} & \textbf{0.47} \\
    & RD & 50.22 & 0.79 & 90.48 & 0.51 & 22.93 & 0.93 & 96.15 & 0.51 \\
\hline
\end{tabular}
}
\label{tab:main_result}

\vspace{-2mm} % Adjust the vertical space after the table as needed

\end{table}
% % % % % % % % % % % % % % % % % % % % % % % % % 

We next addressed the second question: \textbf{\textit{Does SERT improve the model's ability to learn from the teacher model’s reasoning paths?}} Fine-tuning the SERT-activated models with the teacher’s reasoning paths (CoT) revealed several key improvements. \textbf{Improved Task Accuracy:} SERT+RD models achieved higher task accuracy, particularly on StrategyQA, indicating better absorption of the teacher model’s reasoning style. \textbf{Better Output Format Alignment:} These models showed improved formatting consistency and reduced errors, enhancing overall stability. \textbf{Lower Text Repetition:} After CoT distillation, SERT-pretrained models exhibited lower repetition rates, producing more diverse and higher-quality reasoning outputs. \textbf{Controlled Reasoning Length:} Directly distilled models, particularly GPT-2 and GPT-2 medium, tended to generate overly long reasoning paths, often leading to hallucinations. In contrast, SERT-activated models maintained more concise and accurate reasoning, resulting in higher evaluation scores. \textbf{Enhanced Reasoning Quality:} Figure 3 illustrates that SERT+RD improves reasoning quality across all dimensions—coherence, relevance, logical consistency, and completeness. The largest gains were observed in smaller models like GPT-2 small and medium, indicating that SERT has a more substantial impact on smaller models, significantly enhancing their reasoning effectiveness compared to models trained with direct distillation alone.

% % % % % % % % % % % % % % % % % % % % % % % % % 

\begin{figure}[t]
\centering\includegraphics[width=0.43\textwidth]{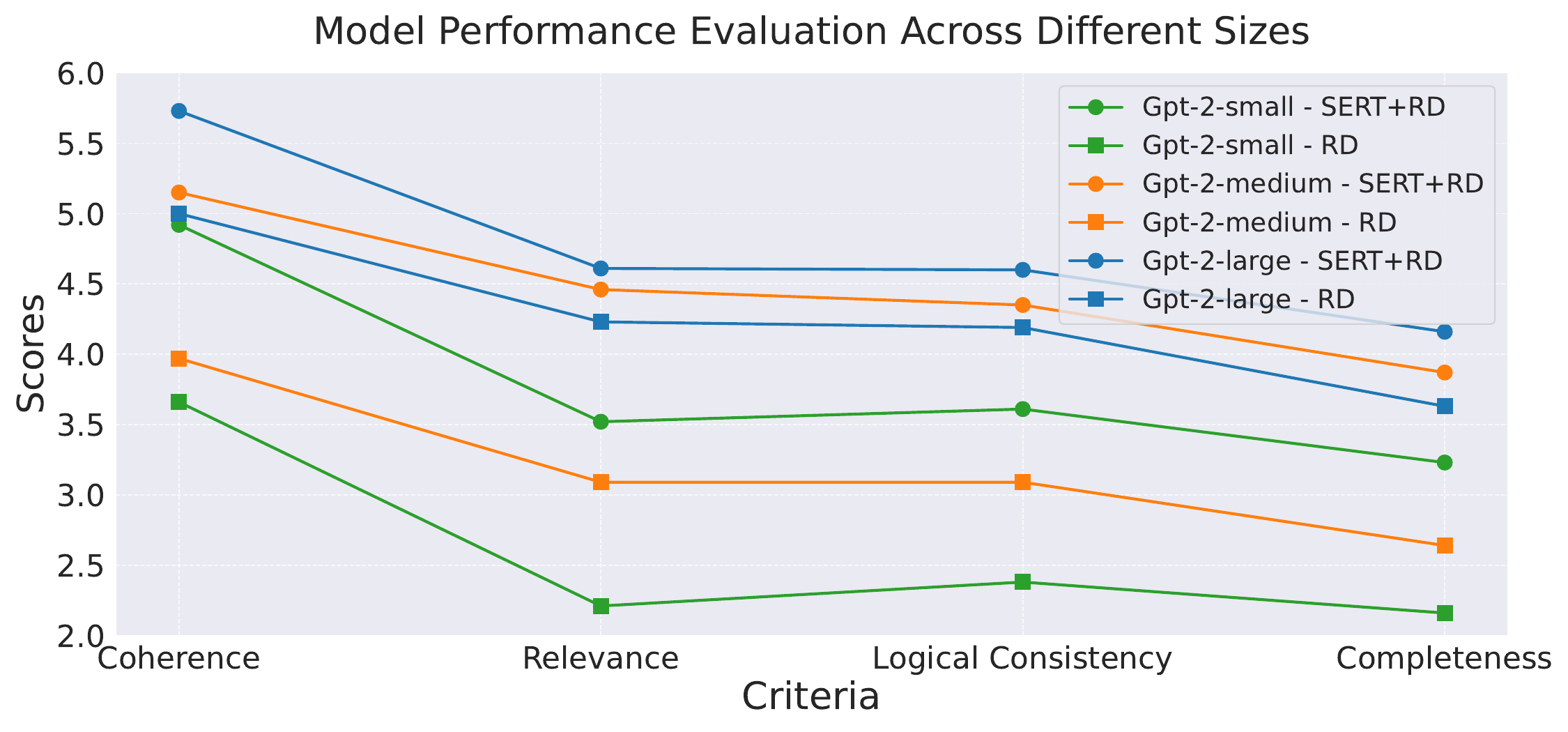}

\vspace{-4mm}%Put here to reduce too much white space after your table 
\caption{Performance comparison of GPT models of different sizes across various evaluation criteria on the CommonSenseQA test set.}
\vspace{-4mm}%Put here to reduce too much white space after your table 

\end{figure}

\begin{figure}[t]
\centering\includegraphics[width=0.49\textwidth]{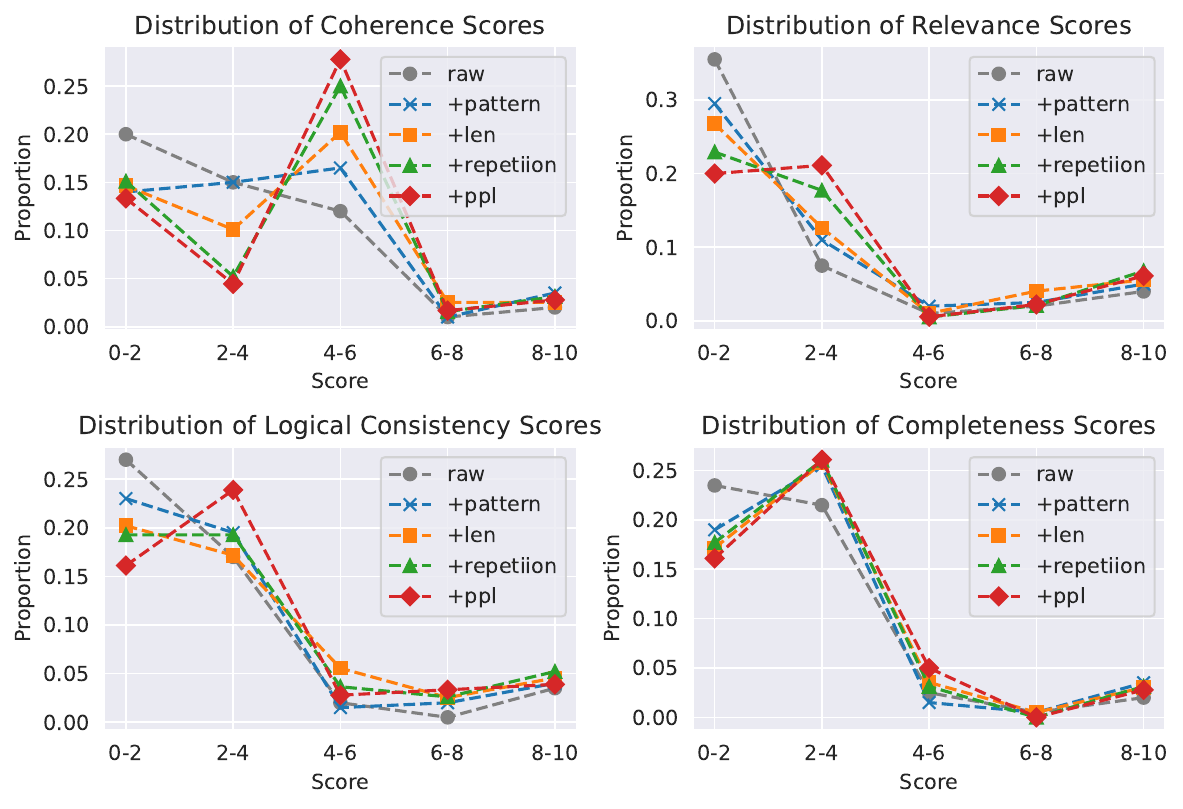}

\vspace{-4mm}%Put here to reduce too much white space after your table 
\caption{Impact of Filtering on Reasoning Quality of CommonsenseQA’s Generated Data}
\vspace{-6mm}%Put here to reduce too much white space after your table 

\end{figure}

% % % % % % % % % % % % % % % % % % % % % % % % % 

\vspace{-2mm}%Put here to reduce too much white space after your table 

\section{Ablation Study}
\vspace{-2mm}%Put here to reduce too much white space after your table 

In this ablation study, we progressively applied each filtering step to assess its impact on the quality of latent reasoning paths generated by the small model. Starting from raw outputs, we sequentially added pattern-based filtering, length filtering, repetition filtering, and finally perplexity filtering (which includes all prior filters). We sampled 100 entries from CommonsenseQA and scored them using Reasoning Detail Evaluation across four dimensions: coherence, relevance, logical consistency, and completeness. Figure 4 shows the distribution of reasoning scores, grouped into score intervals of 0-2, 2-4, 4-6, 6-8, and 8-10.

As shown in Figure 4, the raw model’s outputs are concentrated in the lower score ranges (0-2, 2-4), indicating a high presence of non-reasoning or incomplete responses. As we progressively added filtering conditions, the score distribution shifted towards higher ranges, with the final stage—perplexity filtering, which includes all prior steps—showing the largest improvements across all criteria. The full filtering process resulted in the best performance, significantly reducing low-quality outputs and increasing the proportion of high-scoring responses (6-10).

\vspace{-2mm}%Put here to reduce too much white space after your table 

\section{Conclusion}

% We introduced the Self-Enhanced Reasoning Training (SERT) method, designed to uniquely activate and enhance small models' latent reasoning abilities as demonstrated in zero-shot inference sampling. SERT operates in two stages: it first generates and filters high-quality reasoning paths, and then fine-tunes the model through self-training, significantly increasing the frequency and coherence of reasoning generation. Applied to GPT-2 models on StrategyQA and CommonsenseQA, SERT successfully amplified latent reasoning capabilities, producing higher-quality outputs. When combined with reasoning distillation, SERT-activated models showed marked improvements in reasoning depth and task accuracy, highlighting its effectiveness in both zero-shot reasoning and learning from larger teacher models.

We introduced the Self-Enhanced Reasoning Training (SERT) method, designed to uniquely activate and enhance small models' latent reasoning abilities as demonstrated in zero-shot inference sampling. SERT operates in two stages: first, it generates and filters high-quality reasoning paths, then fine-tunes the model through self-training to boost reasoning frequency and coherence. Applied to GPT-2 models on StrategyQA and CommonsenseQA, SERT enhanced latent reasoning capabilities and produced higher-quality outputs. When combined with reasoning distillation, SERT further improved reasoning depth and task accuracy, highlighting its effectiveness in enhancing both zero-shot reasoning and learning from larger teacher models.

% \textbf{Limitations in Knowledge Transfer During Distillation:}
% Despite these advancements, the results also illuminate a significant limitation in the knowledge transfer capabilities during distillation. Although there is improved alignment, the small models continue to struggle with assimilating new, complex knowledge autonomously, particularly in tasks loaded with domain-specific information such as those in the CommonsenseQA and StrategyQA datasets. While the distillation process bolsters the models' capability to adhere to structured instructions, it does not necessarily equip them with the proficiency to independently generate or comprehend new content beyond the training data provided.

% This observation indicates that while chain-of-thought distillation can refine a model’s ability to mimic and expand on existing reasoning patterns, it does not adequately prepare them for independent knowledge acquisition and application in novel contexts. The small models often depend on replicating learned patterns rather than genuinely understanding and innovating upon the imparted knowledge. This challenge points to an important area for further research into how distillation techniques might be enhanced or supplemented to not only foster alignment in output format but also to significantly improve the genuine understanding and cognitive flexibility of smaller models.

% \vspace{-2mm}%Put here to reduce too much white space after your table 

\clearpage

% In conclusion, Self-Enhanced Reasoning Training (SERT) addresses the critical challenge of distributional mismatches between large teacher models and smaller student models in knowledge distillation. By activating latent reasoning capabilities before distillation, SERT prepares small models for more effective knowledge transfer, effectively narrowing the generation gap. The empirical findings confirm that models activated by SERT and subsequently trained exhibit higher reasoning performance and produce outputs with greater consistency and reduced repetition. These models achieve higher accuracy compared to those trained through direct distillation methods alone, demonstrating SERT's potential as a superior distillation technique for enhancing the reasoning capabilities of smaller models.

\bibliographystyle{IEEEtran} % 使用IEEEtran样式，适用于工程类文章
\bibliography{refs} % 引用你的.bib文件

\end{document}